# A Lightweight Approach to Detection of AI-Generated Texts Using Stylometric Features

Sergey K. Aityan, William Claster, Karthik Sai Emani, Sohni Rais, Thy Tran

Department of Multidisciplinary Engineering, Northeastern University, USA

## Abstract

A growing number of AI-generated texts raise serious concerns. Most existing approaches to AI-generated text detection rely on fine-tuning large transformer models or building ensembles, which are computationally expensive and often provide limited generalization across domains. Existing lightweight alternatives achieved significantly lower accuracy on large datasets. We introduce NEULIF, a lightweight approach that achieves best performance in the lightweight detector class, that does not require extensive computational power and provides high detection accuracy. In our approach, a text is first decomposed into stylometric and readability features which are then used for classification by a compact Convolutional Neural Network (CNN) or Random Forest (RF). Evaluated and tested on the Kaggle AI vs. Human corpus, our models achieve 97% accuracy (≈ 0.95 F1) for CNN and 95% accuracy (≈ 0.94 F1) for the Random Forest, demonstrating high precision and recall, with ROC-AUC scores of 99.5% and 95%, respectively. The CNN (∼ 25 MB) and Random Forest (∼ 10.6 MB) models are orders of magnitude smaller than transformer-based ensembles and can be run efficiently on standard CPU devices, without sacrificing accuracy. This study also highlights the potential of such models for broader applications across languages, domains, and streaming contexts, showing that simplicity, when guided by structural insights, can rival complexity in AI-generated content detection.

## 1 Introduction

Rapid progress in generative AI has taken it to the front lines of modern society and technology. Most major and broadly available generative AI systems including LLMs are continuously trained on vast volumes of available information collecting it from all accessible sources, mostly from the Internet and other datasets. AI can generate content in various formats, including text, voice, music, images, video, software code, many other types of content, and even novel concepts. The generated content makes some slight variations and possibly new features that could be uncommon or even unfeasible in the real world. The artificially generated synthetic content goes to the Internet and is later included in other datasets which are then used for further training of the same AI systems including LLMs. Being trained on the datasets that include AI-generated content, generative AI systems generate new generation of content which is partially based on the knowledge obtained from the training sets that included the generated content. Each AI-generated content may contain little and sometimes unnoticeable variations from the real world, which, in the long run, may accumulate such variations and cause quite significant deviations from the real world similarly to the evolutionary process. In addition, modern LLMs may generate content by making it up, which is quite different from reality. This is known as “AI hallucinations.” Recent studies have shown that the percentage of hallucinated content is quite high among popular LLMs, ranging from 17% to 19% up to 45% of the content [1]. If left without serious attention and the appropriate corrections, AI hallucinations can lead to critical limitations of AI applications that negatively impact human civilization and its progress. AI users may learn from that content that provides deformed and wrong information and representation of the real world. Addressing all kinds of AI hallucinations is a very important and an enormously big task. We decided to start with AI-generated texts and then expand our efforts to other types of AI-generated contents.

The ability to differentiate AI-generated from human-written texts has become increasingly important with the widespread adoption of generative AI. Advanced Large Language Models (LLMs) such as ChatGPT (OpenAI), PaLM (Google), Gemini (Google), Claude (Anthropic), Grok (xAI), DeepSeek (DeepSeek AI) and others can produce fluent, human-like prose [2], making manual identification nearly impossible. Early detection attempts struggled with accuracy. For example, OpenAI's GPT Output Detector achieved only 26% accuracy and misclassified human-written text as AI in 9% of cases before its withdrawal [3]. Recent studies, however, have reported substantially higher detection rates, often in the low 90% range [4–6]. Several online services now offer AI vs. human text classification (e.g., zeroGPT.com), though their reported accuracy remains quite limited.

Research in AI-generated text detection has explored two broad paradigms. Transformer-based classifiers leverage semantic and contextual cues via fine-tuned models such as BERT, RoBERTa, and DistilBERT, can achieve F1 scores and accuracy in the range of 91%-99% in controlled settings [7–12]. However, these models require hundreds of millions of parameters, significantly degrade on unseen domains, and demand substantial computational resources [13]. Variants include zero-shot detectors like Binoculars, which uses perplexity ratios across two LLMs to flag AI text without fine-tuning [14], and ensembles of weak detectors such as Ghostbuster, which aggregate smaller model outputs to improve robustness [15].

Stylometric and hybrid approaches analyze linguistic features such as token statistics, syntactic complexity, readability indices, lexical diversity, and punctuation patterns. Early work demonstrated that simple feature-based classifiers can rival deep models on GPT-2/3 detection [16]. Pure stylometry methods, including Random Forests on 31 stylometric features, achieved up to 98% accuracy [17], while lightweight ensembles combining stylometric, POS, and entropy-based measures reached 85.5% accuracy with minimal computation [18]. Hybrid models that fuse handcrafted features with transformer embeddings, such as RoBERTa+E5 [7] and EssayDetect [19], achieved high accuracy in shared tasks. Approaches like T5LLMCipher leverage embedding clustering for better generalization to unseen models and domains [20]. Comprehensive surveys highlight key challenges, including out-of-distribution detection, adversarial robustness, and evaluation standards, pointing to the need for more generalizable detectors [21].

Most studies are based on the extensive usage of transformers and other heavy AI technologies that require quite significant computing power.

**Gap & Contribution.** Despite these advances, there remains a need for a lightweight solution that (a) exploits rich stylometric descriptors, (b) matches the performance of large transformer ensembles, and (c) requires less computational resources and operates efficiently on standard CPU hardware. Our work addresses this gap with NEULIF, a framework combining a compact Convolutional Neural Network (CNN) and a Random Forest, trained on ~68 stylometric and readability features curated via the 'TextDescriptives' library [22]. On the Kaggle AI-vs-Human dataset, NEULIF achieves 97% accuracy, matching or exceeding heavyweight transformer ensembles, while using less than 105 parameters. We provide a reproducible pipeline, extensive evaluation (accuracy, F1, ROC-AUC, confusion matrices), and discuss deployment considerations, including potential vulnerabilities to adversarial perturbations and strategies for continuous feature updates to track evolving LLM outputs [21]. This approach demonstrates that simplicity, when grounded in linguistic and stylometric insights, can rival complexity in AI-generated text detection.

# 2 Vision and Architecture

## 2.1 Solution vision:

We frame the task as a binary text classification problem given an input text; the model must decide whether the text was generated by an AI system or written by a human. Concretely, we extract stylometric feature vectors from the text and train two separate classifiers: a convolutional neural network (CNN) and a Random Forest. Each model outputs the likelihood of the AI-generated class, after which a threshold is applied to yield one of two labels (AI-generated vs. human-written). By evaluating these models independently, we can assess the effectiveness of lightweight, feature-driven approaches for AI-text detection, highlighting trade-offs between predictive performance and computational efficiency. This binary framing aligns with recent benchmarks [9–12] in AI-text detection and demonstrates that carefully designed stylometric classifiers can 3 provide practical, high-performance alternatives to more complex transformer-based systems.

### 2.1.1 Variable-Length Text Input and Sequence Models

Natural language text is inherently a variable-length sequential input, since different texts may consist of different numbers of characters, words and sentences. Traditionally, processing of such data requires models that can handle input sequences of arbitrary length. Typically, recurrent neural networks (RNNs) and their gated variants like LSTM or GRU are used to process such inputs one element (word or token) at a time and maintain an internal hidden state that captures past context. This design makes RNNs "adept at handling sequences, because they process inputs sequentially and maintain a state reflecting past information" [23]. Similarly, more recent transformer-based models (e.g. BERT, GPT, etc.) rely on self-attention mechanisms that consider relationships among all tokens in the sequence [24]. Indeed, the Transformer architecture excels at processing sequential data by using self-attention mechanisms that model dependencies across entire sequences without recurrence [24]. Both RNNs and transformers naturally accommodate variable-length inputs and capture complex contextual dependencies in text.

However, both RNNs and transformer-based models are computationally intensive, especially when applied to long-form documents. They typically operate on raw token sequences, require extensive training resources, and often involve millions or even billions of parameters. In contrast, our research adopts a lightweight and feature driven approach to the problem of AI-generated text detection. In our approach, we first extract a comprehensive set of predefined linguistic features using efficient NLP pipelines rather than directly relying on deep neural architectures that learn representations directly from raw text. This design dramatically reduces computational overhead, provides good performance and offers a viable and interpretable alternative to the traditionally used resource-heavy sequence models.

### 2.1.2 Linguistic Feature Extraction using spaCy and TextDescriptives

In our implementation, we first convert each text into a fixed-size feature vector, rather than feeding a raw variable length text into a deep word sequence model. We leverage the 'spaCy' library to process raw text and extract rich linguistic annotations [25]. For example, spaCy' tokenizes the text, identifies part-of-speech tags, syntactic dependencies, named entities, and other token-level attributes, producing an object with numerous annotations. In parallel, we use the 'TextDescriptives' tool (a 'spaCy' pipeline component) [22] to compute document-level linguistic metrics using descriptive statistics (e.g. word and sentence counts), readability indices, and measures of syntactic complexity. All of the extracted features are combined into a fixed-length numeric vector for each text. With this feature-engineering approach, the model operates on a compact representation of the text's linguistic profile. In result of this conversion, a variable length text is represented by a fixed-size feature vector, allowing subsequent network layers to process a uniform input shape for all text samples. These sidesteps significantly reduce the computational cost of sequence models and leverages expert-crafted textual features.

Figure 1 illustrates the full pipeline from raw text to final classification. The input text goes through feature extraction process that results its mapping to a fixed feature vector, which is directed to a relatively light convolutional neural network (CNN) or to a random forest (RF) for the final classification.

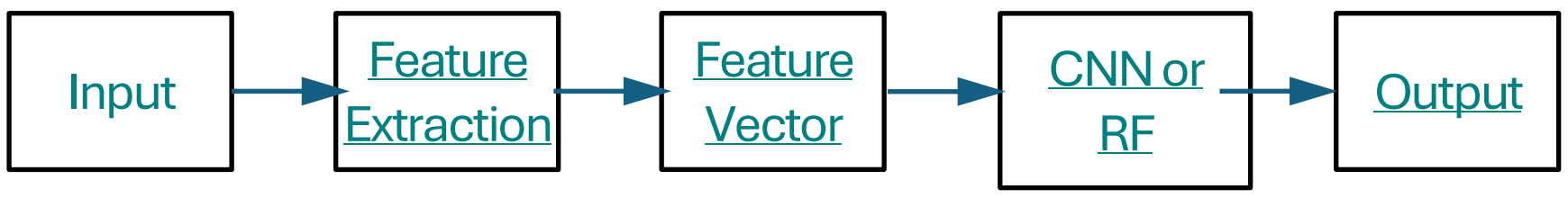

Figure 1: The full pipeline from raw text to final classification

## 2.2 Dataset and Training

We used Kaggle's publicly available "AI vs Human Text" dataset (~500,000 essays). From this corpus, we randomly sampled 20,000 texts, balanced by class (10,000 AI generated vs. 10,000 human-written). Each sample consists of the essay text and a binary label with value 0 for human-written and 1 for AI-generated texts. This binary labeling defines our classification task (AI-generated vs. human-written text).

### 2.2.1 Feature Extraction

We processed each text sample using 'spaCy' (with the 'en core web sm' English model) alongside the 'TextDescriptives' pipeline [22] to extract document-level linguistic features. In total, 68 features per text were obtained, covering multiple categories that capture different aspects of textual style and complexity:

- **Descriptive statistics:** Total token count, number of unique tokens, mean and median token and sentence lengths, and related measures that quantify text length and distribution.
- **Readability indices:** Metrics such as Flesch Reading Ease, Flesch–Kincaid grade level, and Automated Readability Index (ARI), which assess the ease of comprehension.
- **Syntactic and dependency features:** Mean and variance of dependency distances, proportions of part-of-speech (POS) tags (nouns, verbs, adjectives, etc.), and other structural measures.
- **Lexical diversity and richness:** Type-token ratio, entropy of token distributions, and vocabulary variation measures that reflect stylistic diversity.
- **Cohesion and discourse metrics:** Counts of connectives, co-reference chains, and other indicators of text coherence.
- **Complexity and text-quality heuristics:** Sentence and clause ratios, mean parse tree depth, spelling/grammar errors, and sentence variation.

These features were automatically computed by 'TextDescriptives' via spaCy's NLP pipeline, providing a comprehensive representation of each text's linguistic style, which is critical for distinguishing AI-generated from human-written content.

### 2.2.2 Data Splitting and Preprocessing

After feature extraction, some feature vectors contained NaN (Not a Number) or undefined values. These rows with missing values were removed from the dataset. The cleaned dataset was then split in two parts via stratified random sampling to preserve the 50/50 AI/human ratio in each subset. The final partitions were:

- **Training set:** 15,981 samples (80% of data)
- **Validation set:** 1,993 samples (10%)
- **Test set:** 1,997 samples (10%)

All splits contained equal proportions of each class by design. Prior to modeling, we applied standard preprocessing to the features. Specifically, we used scikit-learn's 'StandardScaler' to standardize each feature to zero mean and unit variance. This normalization helps most machine learning models converge more reliably. Finally, any remaining samples with missing values were dropped to ensure data integrity.

# 3 Text Classification with Neural Network

Building on our feature-driven approach outlined in Section 2, we now turn to the neural network architecture that processes the extracted linguistic features. Rather than handling raw text sequences, our model operates on the fixed-size feature vectors derived from 'spaCy' and 'TextDescriptives' [22], enabling efficient computation while maintaining high classification accuracy.

## 3.1 Neural network architecture

Our neural network design reflects a deliberate departure from conventional text processing approaches. While most contemporary methods rely on transformer architectures that process token sequences directly [26, 27], we leverage the rich linguistic features already extracted in our preprocessing pipeline. This design choice offers several advantages: computational efficiency, interpretability, and the ability to capture expert-crafted linguistic patterns that distinguish AI-generated from human-written text.

The architecture processes 68-dimensional linguistic feature vectors through a carefully designed sequence of operations. We employ a convolutional layer to identify local patterns among the features [28, 29], followed by three dense layers that learn complex interactions between different linguistic characteristics. This hybrid approach combines the pattern recognition capabilities of convolutional neural networks with the representational power of fully connected architectures.

**Architecture Overview**

The network flows through four distinct stages, each serving a specific purpose in the classification pipeline. The neural network output neuron fires with the sigmoid activation function for binary classification. The sequential flow from input through convolutional processing, normalization, flattening, and dense layers with regularization enables effective feature learning and classification performance.

**Input Processing**: The model accepts 68-dimensional feature vectors, where each dimension represents a specific linguistic characteristic extracted by our preprocessing pipeline. These features encompass readability metrics, syntactic complexity measures, token-level statistics, and dependency relationship patterns.

**Convolutional Feature Learning**: A convolutional layer with 128 filters and kernel size 3 captures local relationships among neighboring features. This layer identifies patterns that might indicate systematic differences in how AI systems and humans structure their writing. Batch normalization following convolutional operation ensures training stability and faster convergence [30].

**Dense Network Processing**: After flattening the convolutional layer output, three successive dense layers (256, 128, and 64 neurons) progressively refine the feature representations. Each layer uses ReLU activation to introduce non-linearity [31], while dropout regularization (rates of 0.4, 0.3, and 0.2 respectively) prevents overfitting by randomly deactivating neurons during training [32].

**Classification Output**: The final layer employs a single neuron with sigmoid activation, producing a probability score between 0 and 1. This score represents the model's confidence that the input text was generated by an AI system.

The complete architectural specifications, including input/output shapes, parameter counts, and activation functions for each layer, are summarized in Table 1.

Table 1: Neural network architecture

| Layer | Type | Input Shape | Output Shape | Parameters | Activation | Additional |
|---|---|---|---|---|---|---|
| Input | Input Layer | (None, 68,1) | (None, 68,1) | 0 | - | Feature vector input |
| Conv1D | Convolutional | (None, 68,1) | (None, 66,128) | 512 | ReLU | Filters:128, Kernel:3 |
| BatchNorm | Normalization | (None, 66,128) | (None, 66,128) | 512 | - | Stabilizes training |
| Flatten | Reshape | (None, 66,128) | (None, 8448) | 0 | - | Converts to 1D |
| Dense1 | Fully Connected | (None, 8448) | (None, 256) | 2,162,944 | ReLU | Dropout: 0.4 |
| Dense2 | Fully Connected | (None, 256) | (None, 128) | 32,896 | ReLU | Dropout: 0.3 |
| Dense3 | Fully Connected | (None, 128) | (None, 64) | 8,256 | ReLU | Dropout: 0.2 |
| Output | Classification | (None, 64) | (None, 1) | 65 | Sigmoid | Binary output |
| **Total number of trainable parameters**: 2,205,185 (trainable) | | | | | | |

### 3.2 Neural Network Implementation

The model starts with a one-dimensional convolutional layer consisting of 128 kernels of size 3. This layer captures local patterns and relationships among the 68 extracted features. Batch normalization was applied following the convolutional operation to stabilize and accelerate training by normalizing activations.

Subsequently, the output of the convolutional layer was flattened and passed through a series of fully connected (dense) layers. These included three hidden layers with 256, 128, and 64 neurons, each activated using the ReLU function. Dropout technique with rates of 0.4, 0.3, and 0.2 was interleaved between dense layers to prevent overfitting by randomly deactivating a fraction of neurons during training.

The final output layer was a single neuron with a sigmoid activation function, which generates a probability of the model's confidence in whether a given text was generated by an AI or written by a human.

### 3.3 Classification Process and Implementation

The practical implementation of the trained model follows a systematic pipeline that processes raw text input through several transformation stages before producing a final classification decision. The process begins with comprehensive feature extraction, where each input text undergoes analysis through the 'spaCy-TextDescriptives' pipeline [22] to derive a 68-dimensional feature vector. This extraction encompasses multiple linguistic dimensions, including readability indices such as Flesch Reading Ease and Automated Readability Index, syntactic complexity measures including dependency tree depth and sentence structure statistics, lexical diversity metrics capturing vocabulary richness and token length distributions, dependency relation statistics measuring average distances and relation type frequencies, and stylometric features encompassing function word ratios, punctuation density, and part-of-speech distributions.

The extracted features undergo alignment and normalization to ensure compatibility with the trained model's expectations. The raw feature vector is standardized using the fitted 'StandardScaler' transformation, which applies the learned mean and standard deviation from the training data to each feature dimension. This normalization ensures that all features maintain zero mean and unit variance, matching the distribution properties expected by the neural network architecture.

The normalized feature vector requires reshaping to meet the input requirements of the Conv1D neural network. The one-dimensional feature array is transformed into a three-dimensional tensor format, where the first dimension represents the batch size for single sample processing, the second dimension corresponds to the feature sequence length of 68 linguistic features, and the third dimension indicates the feature channels for univariate features.

The trained neural network processes the reshaped input through its convolutional and dense layers, ultimately producing a probability score through the sigmoid activation function in the output layer. This probability represents the model's confidence that the input text was generated by an AI system, with values ranging from 0 to 1. The final classification decision applies a threshold, typically set at 0.5, where probabilities above this threshold indicate AI-generated text and probabilities below indicate human-written content.

This classification pipeline enables real-time processing of text samples with typical inference times under 100 milliseconds on standard CPU hardware. The complete process can be conceptualized as a composite function that sequentially applies feature extraction, standardization, reshaping, neural network inference, and threshold-based decision making. This implementation makes the system efficiently suitable for practical deployment scenarios requiring immediate AI text detection capabilities while maintaining the high accuracy levels.

## 3.4 Test Results

**Evaluation Methodology**

The trained neural network model was evaluated on the test set comprising 1,997 samples (10% of the total dataset) that were kept during training and validation phases. The evaluation employed accuracy as the primary metric to assess model performance in binary classification.

**Metrics and Evaluation**

The model's performance was assessed using accuracy as the primary measure, rep resenting the proportion of correctly classified samples. The model achieved 97% accuracy on the held-out test set, indicating that approximately 97 out of every 100 text samples were correctly classified on the test set.

Accuracy serves as the primary measure of the overall model performance, representing the proportion of correctly classified samples. This high accuracy demonstrates the model's exceptional ability to distinguish between AI-generated and human-written text across the diverse test dataset. The accuracy results reflect the model's consistent performance in correctly identifying both classes without significant bias toward either AI-generated or human written content as indicated by ROC-AUC in Section 5.1. The model correctly classified 1,932 instances from 1,997 test samples with only 65 misclassifications across both categories. This level of accuracy represents a substantial achievement in the challenging domain of AI text detection, where subtle linguistic differences must be captured to differentiate between synthetic and authentic content.

To complement the accuracy assessment, additional metrics confirm the model's robust classification capabilities. The F1-score of 97% provides a harmonious balance between precision and recall, representing the harmonic mean of these two critical metrics. Precision of 97% indicates that when the model predicts a text as AI-generated, it is correct 97% of the time, while recall of 96% shows that the model successfully identifies 96 percent of all actual AI-generated texts. The ROC-AUC of 99.51% is particularly noteworthy, indicating an almost perfect ability to distinguish between the two classes across various classification thresholds. This metric reveals the model's exceptional discriminative power, suggesting that at nearly any decision boundary, the neural network can effectively separate AI-generated from human-written texts. Moreover, the low logarithmic loss (0.0915) provides critical insight into the model's probabilistic predictions. This extremely low log loss confirms that the model's confidence scores are exceptionally well-calibrated, meaning the predicted probabilities closely reflect the actual likelihood of a text being AI-generated or human-written.

The neural network demonstrated 97% accuracy, establishing its effectiveness in distinguishing between AI-generated and human-written text samples. This accuracy level translates to successful classification of nearly 1,932 out of 1,997 test instances.

The model's performance profile confirms robust classification capabilities across multiple evaluation dimensions. These metrics collectively validate the effectiveness of the stylometric feature extraction approach combined with the convolutional neural network architecture.

This performance profile confirms that the stylometric feature-based neural network architecture delivers competitive results matching or surpassing existing literature benchmarks, while offering computational advantages for practical deployment scenarios requiring real-time text classification capabilities.

**Confusion Matrix Analysis**

The confusion matrix analysis presented in Figure 2 provides a detailed breakdown of the classification performance of the model in different categories. The chart illustrates the distribution of classification outcomes for the test dataset, revealing four key performance indicators.

**True Negative (969 samples)**: Human-written texts correctly identified as human-written, representing 48.5% of the total test set. This high count demonstrates the model's strong capability in preserving the integrity of authentic human writing, with minimal false accusations of AI generation.

**True Positive (963 samples)**: AI-generated texts correctly identified as AI generated, comprising 48.2% of the test dataset. This substantial number indicates excellent detection capability for AI-generated content, successfully flagging nearly all synthetic text samples.

**False Negative (37 samples)**: AI-generated texts incorrectly classified as human written, accounting for only 1.9% of the total samples. This low error rate suggests that the model rarely fails to detect AI-generated content, maintaining high sensitivity for synthetic text identification.

**False Positive (28 samples)**: Human-written texts incorrectly classified as AI generated, representing merely 1.4% of the test set. This minimal misclassification rate indicates that the model maintains excellent specificity, rarely misidentifying authentic human writing as AI-generated.

**Total Samples (1,997):** The complete test dataset used for final evaluation, rep resenting 10% of the original dataset and providing a robust foundation for the performance assessment.

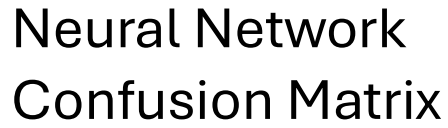

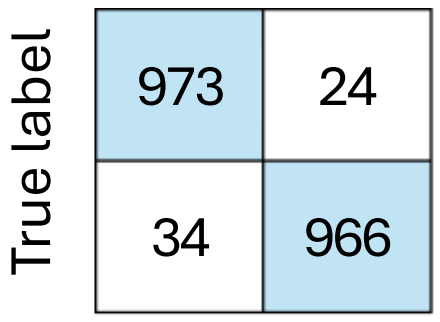

Figure 2: Confusion matrix for recognition of AI-generated vs human-written texts by the neural network

The confusion matrix results confirm the model's exceptional performance with a near-perfect balance between true positives and true negatives (963 vs. 969), while maintaining remarkably low error rates for both false classifications. The slight predominance of false negatives over false positives (37 vs. 28) suggests that the model exhibits a conservative bias, occasionally allowing AI-generated text to pass undetected rather than incorrectly flagging human writing. This characteristic may be preferable in practical applications where preserving human authorship integrity is prioritized over perfect AI detection.

## 4 Text Classification with Random Forest

### 4.1 Random Forest Architecture

The Random Forest architecture demonstrates enhanced performance in AI-generated text detection when integrated with 'TextDescriptives' stylometric feature processing. This combination leverages 'TextDescriptives' comprehensive suite of text statis tics—including readability indices, syntactic complexity metrics, and lexical diversity measures—to capture nuanced linguistic properties. By representing a broader spectrum of textual characteristics beyond surface-level patterns, these features significantly improve the model's discriminative capability for identifying synthetic content [33].

**Architecture Overview**

Random Forest is a collection of many decision trees. Each tree is trained independently on a slightly different subset of the training data. The architecture consists of the following components as shown in Figure 3.

**Decision trees**: The larger number of trees generally leads to better performance, as it reduces the variance and improves the robustness of the model. However, it also increases the training time and memory usage. Each tree being trained with the respective random subset dataset, and then at prediction time, every tree independently outputs a class label for the input sample.

**Ensemble Aggregation**: Final prediction is the majority voting for all predictions across all trees.

**Randomness Mechanisms**: The randomness is achieved by random feature subsets per split decorrelate trees, and bootstrap sampling ensures each tree sees different data.

A spaCy model ('en core web sm'): This is a pretrained statistical model for the English language provided by 'spaCy'. It's the foundation for understanding text data. It was pre-trained on a massive dataset of English text. It has learned patterns and relationships between words, grammar, and context. When loading this model, it allows access to a wide range of natural language processing capabilities.

**A 'textdescriptives/all' pipeline**: This is a custom pipeline component that being added to the 'spaCy' model. It's provided by the 'TextDescriptives' library, and its purpose is to extract a variety of descriptive features from the text. When added, it automates the process of extracting potentially useful features that can be used as an input to the machine learning model.

**An instance of the RandomForestClassifier class with hyperparameter**: This object will be the machine learning model used for classification tasks. Scikit learn's 'RandomForestClassifier' is a powerful and versatile ensemble learning method based on decision trees, widely used for its robustness, ability to handle high dimensional data, and inherent feature importance estimation, making it well-suited for text classification. During training, the model learns the relationships between the features of the text (e.g., readability scores, lexical diversity, word usage patterns) and the class label (AI or Human). After training, the model can predict the class label of new, unseen text. Finally, it combines the predictions of multiple decision trees where each tree is trained on a random subset of the data and a random subset of the features. The 'RandomForestClassifier' achieves this by combining the predictions of multiple decision trees, where each tree is trained on a random subset of the data and a random subset of the features. This ensemble approach reduces overfitting, improves generalization, and provides a more stable and reliable prediction than individual decision trees. These hyperparameters serve different purposes as follows:

- n_estimators: specifying the number of decision trees in the forest.
- criterion: specifies the function used to measure the quality of a split
- random state: specifies the seed for the random number generator.

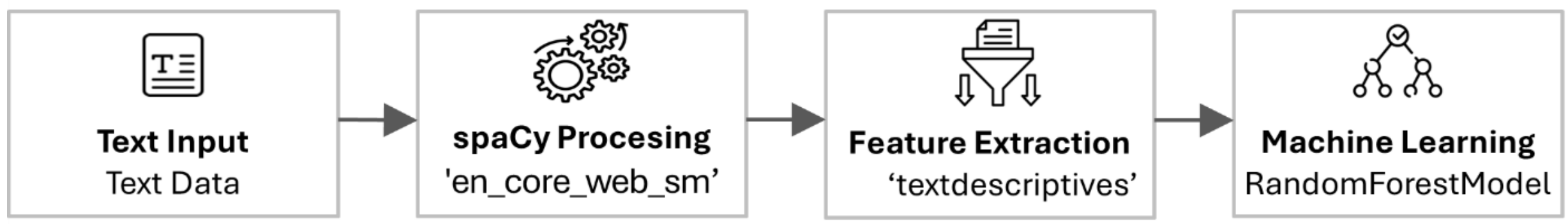

Figure 3: Random Forest Solution Architecture Diagram

The effectiveness of Random Forest Classifier in 'scikit-learn' stems from its specific architecture and the way how it combines multiple decision trees. To further understand how hyperparameter tuning impacts performance, it's crucial to examine the internal workings of the Random Forest. The architecture of a Random Forest Classifier is illustrated using Figure 4.

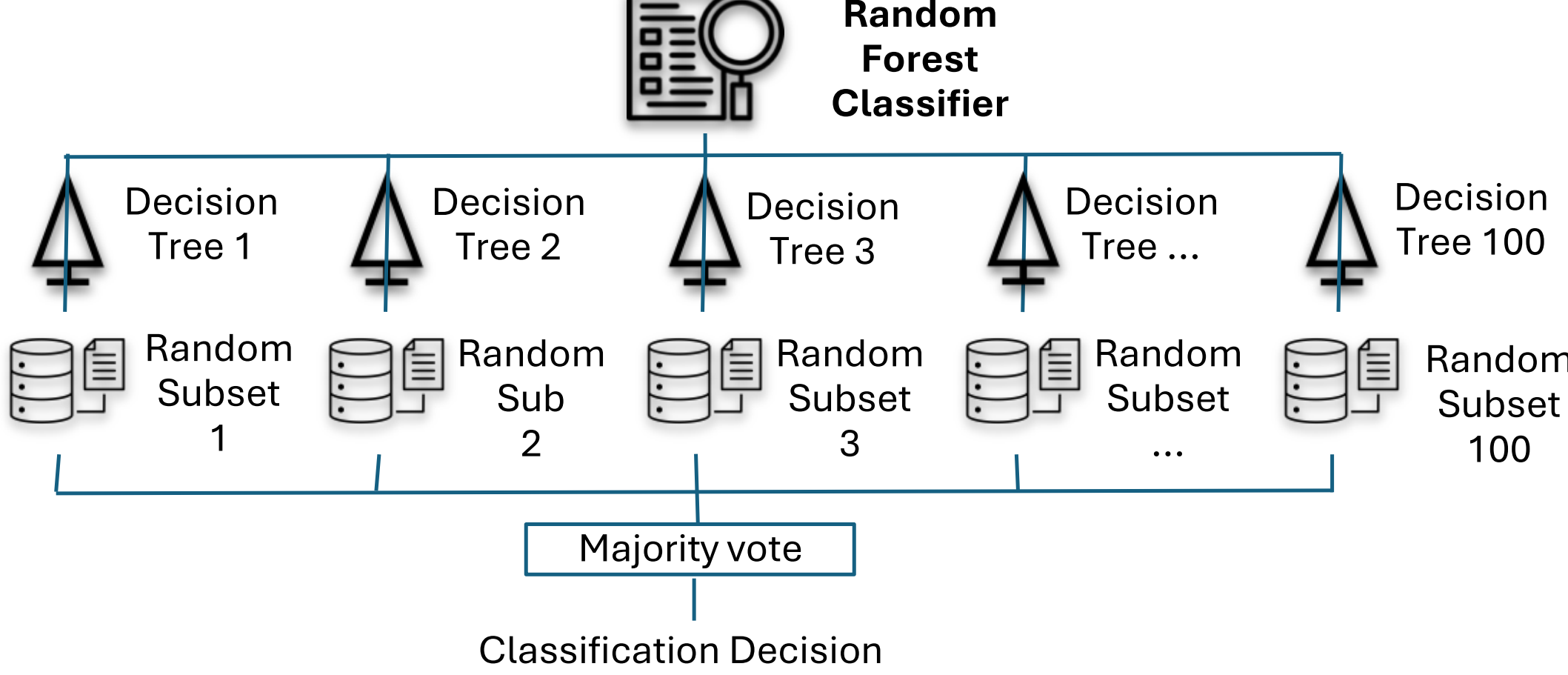

Figure 4: Random Forest Classifier Diagram

## 4.2 Random Forest Implementation

The environment and development processes were implemented in a Python/Jupyter environment as defined in Section 3. Linguistic feature extraction was performed using 'spacy' with 'TextDescriptives'. The data pipeline, described earlier, managed dataset acquisition from 'Kaggle', data loading, NLP processing with 'spacy' and 'TextDescriptives', and feature extraction. For training, we used 'RandomForestClassifier', constructed 100 decision trees (n estimators=100) with the Gini impurity criterion for splitting nodes, and random state=42 to set the seed for the random number generator.

## 4.3 Classification Process and Implementation

The classification process aims to build a model capable of accurately distinguishing between AI-generated and human-written text. It begins with data acquisition and preparation, involving downloading a relevant dataset and balancing the classes to prevent bias. Next, the text data undergoes feature extraction, leveraging 'spaCy' and 'TextDescriptives' to derive meaningful linguistic features. The prepared data is then partitioned into training, validation, and test sets to facilitate model training, hyper parameter tuning, and unbiased performance evaluation. A machine learning model, specifically a 'RandomForestClassifier,' is selected and trained on the training data. Hyperparameter tuning, often employing techniques like 'GridSearchCV', is performed to optimize the model's performance on the validation set. Finally, the trained model's generalization ability is rigorously assessed using the held-out test set, providing an estimate of its performance on unseen data. Key performance metrics, such as accuracy, precision, recall, and F1-score, are monitored throughout the process to ensure the model meets the desired performance criteria.

We implemented a process that consists of several steps required to build and evaluate the AI-generated vs. Human-written text classifier. It starts with using 'kagglehub' to download the dataset and then balances the dataset to ensure equal representation of AI-generated and human-written text. The code then leverages 'spacy', enhanced with the 'TextDescriptives' pipeline, to extract a rich set of linguistic features from the text data. The 'train test split' function from 'scikit-learn' is used to partition the data into training, validation, and test sets, ensuring proper data segregation for model development and evaluation. A 'RandomForestClassifier' is instantiated and trained using the training data. We employed 'GridSearchCV' to optimize the model's hyperparameters by conducting systematic search for the best combination of parameters based on cross-validation performance. The trained model's performance is then evaluated on the validation and test sets using metrics like accuracy, classification reports, and confusion matrices to provide a comprehensive assessment of its capabilities. Finally, the trained model is serialized and saved using 'joblib' for subsequent deployment and reuse, eliminating the need for retraining. The implementation also includes performance benchmarking using 'cProfile', memory profiler, and 'psutil' to identify potential bottlenecks and optimize resource usage.

## 4.4 Results

### Evaluation Methodology

Random Forest Solution and Convolutional Neural Network Solution models were trained and evaluated on identical datasets, enabling direct performance comparison as validated in AI-generated text detection.

### Metrics and Evaluation

The model's performance was evaluated using a comprehensive suite of complementary metrics, ensuring a thorough assessment of its classification capabilities. Standard binary classification metrics were used to evaluate both class-specific and overall performance.

Consistent with the previously described evaluation strategy, accuracy served as the primary measure of overall model performance. The model achieved 94% accuracy on the held-out test set, correctly classifying on average 94 out of every 100 text samples.

To assess class-specific performance, precision and recall were calculated. The model demonstrated a precision of 94% for AI-generated text, indicating a low rate of false positives. The recall of 95% showed high sensitivity in detecting AI-generated content. The F1-score, which balances precision and recall, was 94%.

The Area Under the Receiver Operating Characteristic Curve (ROC-AUC) was used to evaluate the model's ability to discriminate between AI-generated and human written text across various classification thresholds. The ROC-AUC of 95% indicated near-perfect separation between the two classes, demonstrating robust performance irrespective of threshold selection.

Finally, logarithmic loss was used to assess the quality of the model's probabilistic predictions. The achieved log loss of 1.6039 suggests well-calibrated probability estimates, providing meaningful confidence scores for classification decisions.

**Confusion Matrix Analysis**

Confusion Matrix visualizes and quantifies how well a model classifies instances cor rectly by comparing predicted values with actual ground truths. It helps machine learning models improve by exposing their weaknesses in prediction accuracy, preci sion, and recall. The test confusion matrix evaluates final performance on the data the model has never seen, provided an unbiased estimate. of how the model performs on novel data, and validates whether improvements during tuning generalized well. The test confusion matrix is illustrated inof how the model performs on novel data, and validates whether improvements during tuning generalized well. The test confusion matrix is illustrated in Figure 5.

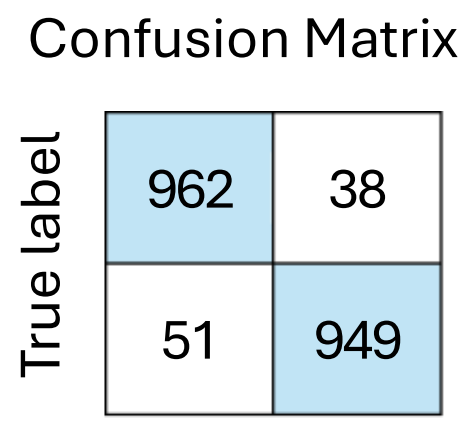

Figure 5: Confusion matrix for recognition of AI-generated vs human-written texts by Random Forest

**True Negative (962 samples)**: The number of samples that were negative and the model correctly predicted them as negative.

**True Positive (949 samples)**: The number of samples that were positive (e.g., had the condition, were fraudulent, etc.) and the model correctly predicted them as positive.

**False Negative (38 samples)**: The number of samples that were positive, but the model incorrectly predicted them as negative.

**False Positive (51 samples)**: The number of samples that were negative, but the model incorrectly predicted them as positive.

**Total Samples (2000)**: The total number of samples or data points tested.

The observed difference in sample counts between the two approaches arises from their respective preprocessing methods: the neural network requires explicit 3D inputs for Conv1D (resulting in 1,997 samples after handling NaN values), whereas the random forest uses a TF-IDF sparse matrix, which inherently excludes empty rows.

**Performance Validation**

The above results suggest that the model is not only highly accurate, but also reliable in both identifying positive cases and minimizing false alarms. The low numbers of false positives and false negatives further reinforce the model's robustness and suitability for deployment in real-world scenarios. To ensure these results are generalizable, the validation process was performed on a dataset independent from training, providing confidence that the model will perform similarly on new, unseen data. Regular monitoring and potential recalibration are recommended to maintain this high performance over time.

**Accuracy**

Validation accuracy guides hyperparameter tuning to prevent overfitting, while test accuracy provides a final, unbiased performance estimate on unseen data. Ideally, both scores should be high and similar, indicating a robust and generalizable model; significant discrepancies suggest overfitting or a non-representative validation set.

**Validation Accuracy** is equal to 0.94 (94%). This is calculated on a portion of the data (the validation set) that the model does not see during training but is used during model development for tuning hyperparameters and evaluating model performance before the final test.

**Test Accuracy** is equal to 0.96 (96%). This is calculated on the completely unseen test set, which the model never encounters during training or validation. This provides a final, unbiased estimate of how the model will perform in real-world scenarios.

# 5 Analysis of Results

## 5.1 Analysis of Classification Accuracy

The accuracy achieved with our solution is competitive with results published by other research groups [9–12], who primarily used computationally demanding transformer-based solutions. As shown in Table 4, our Neural Network and Random Forest models achieve accuracies of 97% and 95%, respectively, along with high F1-scores, precision, recall, ROC-AUC, and low log loss. We discuss the benchmark against other solutions in section 5.2.

Table 2: Comparison of performance metrics for Neural Network and Random Forest approaches. Metrics reported include accuracy, F1-score, precision, recall, ROC-AUC, and log loss.

| Metric | Neural Network | Random Forest | Interpretation |
|---|---|---|---|
| Accuracy | 97% | 95% | Highly accurate, correctly classifying the vast majority of cases. |
| F1-Score | 97% | 94% | The model maintains a strong balance between not missing positives and not misclassifying negatives as positives. The slightly lower F1-score suggests a very small imbalance between these two. |
| - Precision | 97% | 95% | The model makes few false positive errors. When it predicts a positive, it's usually right. |
| - Recall | 96% | 94% | The model rarely misses true positives. It catches most of the real cases. |

| ROC-AUC | 99.5% | 95.6% | The model is excellent at distinguishing between the positive and negative classes. |
|---|---|---|---|
| Log Loss | 0.0915% | 1.6039 | The model often predicts correctly, but sometimes not very confident. Needs some improvement. |

**Neural Network Approach**

The neural-network classifier achieves high performance on all metrics (Table 2). Its overall accuracy is 97%, meaning it correctly classified 97% of test samples. This near-perfect accuracy is expected for a model that has both high precision and recall. Precision is 97%, indicating that almost all instances labeled as "AI-generated" were truly AI (very few human texts were misclassified as AI). Recall (true positive rate) is 96%, meaning the model correctly identified 96% of all AI texts (very few AI texts were missed). The high F1-score of 97%– the harmonic mean of precision and recall confirms that precision and recall are both strong and well-balanced, so the model is not skewed toward one class.

- **ROC-AUC (99.5%)**: This measures the model's ability to separate the two classes across all thresholds. An AUC of 0.9951 is extremely close to 1.0, implying almost perfect discrimination. In other words, at nearly any decision threshold the NN distinguishes human vs. AI text very well.
- **Log Loss (0.0915)**: This value indicates that the NN's predicted probabilities are well-calibrated. The model not only classifies correctly but also assigns probabilities that closely reflect true likelihoods. In practice, this low log loss means the NN's confidence estimates are far more reliable, providing a strong probabilistic interpretation of its predictions.
- **Balance and Trade-offs**: The NN's precision (97%) and recall (96%) are nearly identical, indicating a well-balanced decision threshold. Both false positives and false negatives are minimal, reflecting strong overall model symmetry. Its trade-off lies in model interpretability rather than predictive balance.

**Random Forest Approach**

The random-forest (RF) model is also highly accurate, but slightly weaker than the neural networks (NN) model. Its accuracy of 95% means that the model correctly classifies about 95% of texts. The RF's precision and recall are both 95%. This means that RF correctly predicts "AI-generated" texts in 95% cases and captures 95% of all true AI-generated texts. The F1-score of 94% reflects similar levels of precision and recall (slightly lower due to a small class trade-off). These values reflect that the RF makes relatively few mistakes in either direction.

- **ROC-AUC (95.55%)**: The RF's AUC of 0.9555 is still very high, meaning it effectively distinguishes human vs. AI samples across thresholds. However, it is noticeably below the NN's 0.9951, signalling less robust separation.
- **Log Loss (1.6039)**: This is much higher than the NN's. A log loss of 1.60 indicates that many of the RF's probability outputs are not well calibrated, it often predicts classes correctly but with lower confidence. In practice, the RF's higher log loss means its predicted probabilities give less reliable confidence than the NN's, and there is room for calibration improvement.
- **Balance and Trade-offs**: The RF's precision ≈ recall situation (~95% each) indicates it used a symmetric threshold. Both models have very strong balance of false positives vs false negatives, but the RF errs slightly more on human AI mislabels.

Overall, the Random Forest is a robust classifier (accuracy ~ 95%) but it trails the neural network on every metric. Its lower AUC and much higher log loss imply less discriminative power and poorer probability calibration. In addition to performance differences, the Random Forest model occupies

~10.6 MB, whereas the neural network is 25 MB. Despite its larger size, the neural network maintains superior AUC and probability calibration, making it the more powerful classifier for distinguishing AI generated from human text, while still being orders of magnitude smaller than typical transformer-based ensembles.

Figure 6 presents the side-by-side comparison of confusion matrices for the neural network (left) and random forest (right) on the test set. Both models demonstrate strong performance, with high true positive and true negative counts. The neural network achieves a balanced error profile, while the random forest slightly reduces false positives, reflecting a more cautious stance in classifying human-written text.

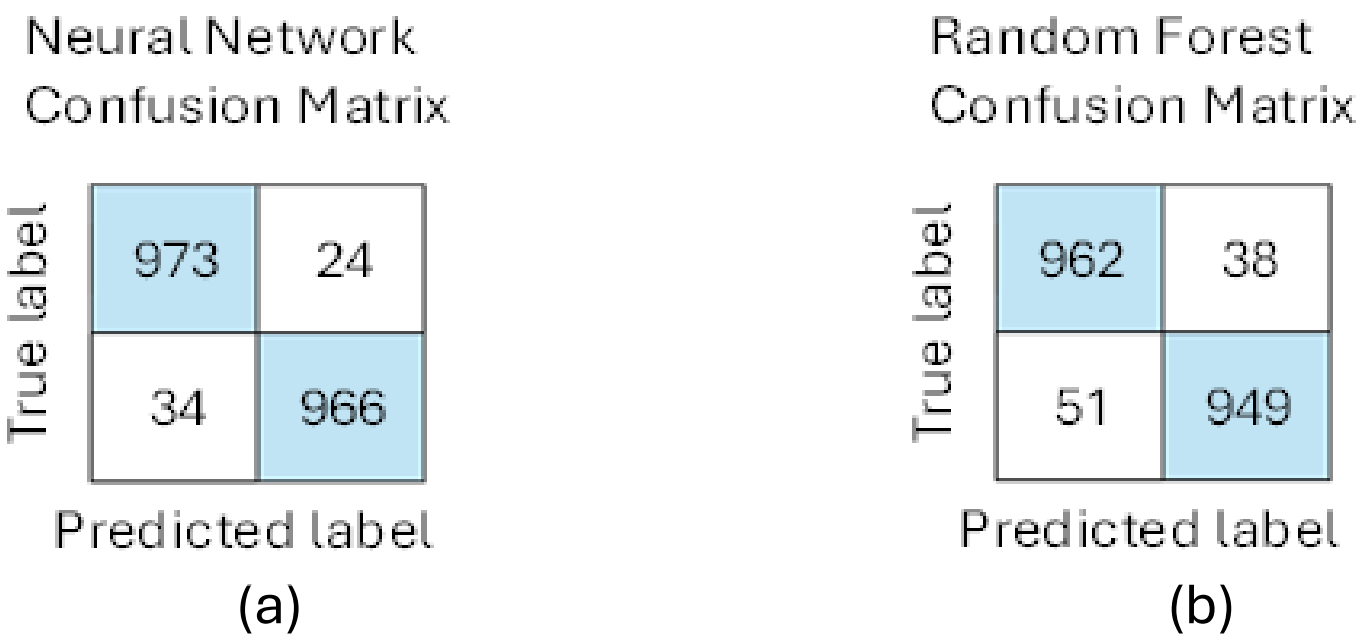

Figure 6: Test confusion matrix of (a) neural network and (b) Random Forest models

## 5.2 Benchmarking with other Existing solutions

Table 3 compares our lightweight models against recent AI-text detectors from the literature. We report test accuracy, along with each method's approach and model class. Notably, our CNN and RF classifiers (trained on fixed-length linguistic features) achieved 97% accuracy.

Table 3: Performance comparison of AI-text detection methods

| Method | Accuracy (%) | Approach | Model Class |
|---|---|---|---|
| AI-generated Text Detection [7] | 99% (F1 = 0.994) | Optimized neural architecture | Transformer-based ensemble |
| DeTeCtive [9] | 96% (AvgRec) | Multi-level contrastive learning | Transformer-based |
| Restricted Embeddings [10] | 91%–95% | Subspace pruning of embeddings | Transformer-based |
| ChatGPT Detector [11] | 98%–99% | Fine-tuned CamemBERTa/XLM-R | Transformer-based |
| RoBERTa + BiLSTM [12] | 97% (open-set) | Fine-tuned RoBERTa + BiLSTM | Transformer + RNN |
| StyloAI [17]* | 81% | Traditional ML classifier and stylometric features | Random Forest based |
| NEULIF (1D CNN) – our current approach | 97% | Linguistic stylometric feature vectors | Lightweight convolutional neural network |
| NEULIF (Random Forest) - our current approach | 95% | Tree-based on linguistic stylometric feature vectors | Traditional machine learning |

*StyloAI reported 81% accuracy on a reasonably large AuTextification dataset (55,677 samples) and 98% accuracy on a quite small Education dataset (less than 200 samples for all including training, validation, and testing that may cause overfitting resulted in artificially elevated accuracy reported by Opara [17].

Among transformer-based approaches, Abburi et al. [7] employed an optimized neural ensemble for binary AI-text classification, achieving 99% F1. "DeTeCtive", [9] used a multi-level contrastive learning framework over large text encoders (e.g., fine-tuned Transformers) to distinguish writing styles. In cross-domain evaluation, DeTeCtive attained ~96%–99% average recall (~accuracy). Similarly, Kuznetsov et al. [10] used BERT/RoBERTa embeddings with subspace pruning for robust detection; they reported large gains in out-of-distribution (OOD) accuracy (improving mean OOD score by up to 9%–14%over baselines), indicating high in-domain performance (around 91% baseline) enhanced via embedding restriction. Antoun et al. [11] evaluated ChatGPT detectors (CamemBERTa/XLM-R classifiers) on French text and observed ~98%–99% in-domain accuracy/F1; however, they cautioned that such detectors dropped substantially on novel or adversarial text (OOD). Petropoulos & Petropoulos [12] used a fine-tuned RoBERTa+BiLSTM architecture to achieve ~97%–99% accuracy and F1 in the shared task evaluation. To summarize the above, transformer-based detectors generally reach high in-domain scores (often ~95%–99%), but at the cost of large "heavy" models that require extensive computing power and poorer generalization.

It is worth noting that StyloAI's accuracy 98% on the Education dataset [17] should be interpreted cautiously because the dataset has only 200 samples (100 AI generated,100 Human written). The 17-percentage points gap between the Education dataset (98%) and the Larger AuTextification dataset (81% on 55,677 samples) suggests potential overfitting to this limited sample set rather than generalized detection capability.

Among lightweight approaches, NEULIF achieves state-of-the-art performance. Our 1D CNN attains 97% accuracy,16 percentage points higher than StyloAI's 81% on large-scale evaluation, while our Random Forest variant achieves 95%. This positions NEULIF as the best-performing lightweight AI-text detector, achieving accuracy competitive with heavyweight transformer-based methods while maintaining orders of magnitude smaller model size (see Section 5.1). These results suggest that linguistic features plus simple classifiers can rival complex deep models as demonstrated on our dataset. Typically, methods are evaluated on held-out test sets that come from the same distribution as the training data (i.e., in-domain). For example, Antoun et al. (2023) noted that their "ChatGPT detector" has 99%+ accuracy in-domain but falls on novel or adversarial texts (OOD). Similarly, our results are reported in-domain only.

## 6 Conclusions

In this paper, we introduced NEULIF, a lightweight framework for AI-text detection that achieves state-of-the-art performance in the lightweight detector class based on fixed-length linguistic features and compact classifiers. Our CNN and Random Forest models achieve accuracies of 97% and 95%, respectively, significantly outperforming existing lightweight methods (see Section 5.2) placing them on par with or exceeding the performance of more computationally intensive transformer-based methods reported in recent literature [9–12]. Despite their simplicity, both models demonstrate robust precision–recall balance, high discriminative ability (ROC-AUC up to 0.995), and well-calibrated probability estimates (log loss 0.0915), highlighting that lightweight architectures can deliver state-of-the-art reliability without the overhead of large-scale encoders.

Compared with transformer-based detectors, which often require substantial computing and memory resources, our approach achieves near-transformer accuracy at a fraction of the computational cost, thereby demonstrating the value of feature-driven, efficient models. This efficiency is particularly advantageous for deployment in resource-constrained environments, such as educational tools, low-power devices, or real-time detection pipelines, where the costs of running large models are prohibitive. Beyond demonstrating strong in-domain performance, our findings reinforce a broader methodological point: compact models, when paired with carefully engineered linguistic features, can remain competitive with deep transformer architectures in accuracy while being orders of magnitude lighter. Our system works with high efficiency and accuracy of AI-generated texts detection on regular laptops and desktop computers. This opens promising directions for the design of sustainable and accessible AI-detection tools. In our future research, having established NEULIF as the best-performing lightweight detector, we will focus on the generalization of these models across domains, languages, and adversarial settings, where lightweight methods may complement or even surpass heavyweight solutions. An additional line of inquiry involves assessing model performance on texts spanning multiple centuries, enabling the study of diachronic language variation and the evolution of stylistic markers over time. Beyond text, a promising direction is the development of similarly lightweight yet accurate approaches for other modalities such as audio, video, and images, extending the principles demonstrated here to broader multimodal detection tasks. In conclusion, our study provides evidence that simplicity, when grounded in domain-specific structural insight, can rival complexity in AI-generated content detection.